\documentclass[letterpaper]{article} 
\usepackage{aaai2027} 
\usepackage[hyphens]{url}  
\usepackage{graphicx} 
\usepackage{natbib}  
\usepackage{caption} 
\usepackage{algorithm}
\usepackage{algorithmic}
\usepackage{booktabs}
\usepackage{amsthm}

\title{Look Before You Leap:\\ Pre-Action Verification for LLM Agents}

\author{
  Asaad Althoubi \\
}
\affiliations{
    Oklahoma State University \\
  Stillwater, OK USA \\
  \texttt{aalthou@okstate.edu}
  }

\nocopyright

\begin{document}

\maketitle

\begin{abstract}
An LLM agent acts on the world by emitting actions: shell commands to run, edits to apply. A wrong action does not always fail loudly; it can fail silently, producing a plausible but incorrect effect that raises no error. We argue that a cheap deterministic check, run before an action takes effect, is an effective and underused form of agent oversight, and we study it across two action modalities in one framework. The idea is to fix an action's correct effect by construction, before any executor runs, so that silent failure is measured directly and the verifier may abstain rather than guess. For shell commands, a static verifier over 9{,}930 commands and 482 tools catches 95.8\% of invalid commands at a 10.0\% false-positive rate. Its syntax and binary checks are oracle-exact, giving zero false positives while catching half of all errors; the flag check is bounded only by help-text coverage and accounts for every false positive. For code edits, a benchmark of 640 edits over 224 files isolating the apply step exposes a sharp split. Content-anchored formats such as search/replace and diff fail cleanly, whereas location-anchored formats fail silently: line numbers corrupt 99.1\% of files under a one-line shift, and function-name edits hit the wrong function 12.7\% of the time. In both settings a refuse-when-unsure policy turns silent failures into recoverable ones at a tunable cost in applicability: selective grounding reaches 0.958 recall at 7.0\% false positives, and an anchor-and-verify applier records one silent misapplication in 8{,}320 trials (0.01\%). We release both benchmarks, the verifiers, and the guards.
\end{abstract}



\begin{links}
    \link{Code}{Will-be-released.}
    \link{Datasets}{Will-be-released.}
\end{links}

\section{Introduction}
Language-model agents increasingly take consequential actions. They run shell commands in a terminal and apply edits to source files \citep{swebench,sweagent,openhands,osworld}. Unlike a chat response, an action changes state, and a wrong action need not announce itself. A command with a hallucinated flag may error out, but it may instead do something subtly different from what was intended. An edit aimed at the wrong location may fail to apply, or it may apply cleanly to the wrong place and corrupt the file with no error at all. We call the second case a \emph{silent failure}: the action appears to succeed, so neither the agent nor a downstream check is alerted, and the error propagates into later steps. When agents act over long, unattended trajectories, one silent action early on can quietly invalidate everything after it, and so the reliability of individual actions underpins the trustworthiness of the whole.

These cases are concrete. An agent that emits \texttt{tar -xf archive.tgz} (omitting \texttt{z}), or that passes a flag the installed version does not support, gets an error message it can read and recover from. An agent that expresses an edit as ``replace lines 40--52'' against a file that has since shifted by one line overwrites the wrong twelve lines and reports success, and the corruption surfaces only much later, if at all. The first failure is loud and recoverable; the second is silent and damaging. The difference is not luck. It follows from how the action is represented.

Silent failure is an oversight problem. A model cannot reliably catch its own mistakes by introspection, and post-hoc repair \citep{selfrefine,selfrepair} triggers only once an error is observed, which a silent failure by definition is not. We study a complementary mechanism: a cheap deterministic check placed before the action takes effect, which either admits the action or refuses it and hands back a recoverable signal. This is oversight by construction rather than by a second, fallible model call. It is model-agnostic, it costs no inference, and it is the kind of safe-by-design, empirically evaluated guard that deployed agents need.

One methodological move makes the problem measurable, applied to two action modalities. For each action we fix the correct effect by construction, before any executor or applier runs. Every trial then falls into one of three classes: a \textbf{success}, where the effect matches the target; a \textbf{clean failure}, where the action is refused or does not apply and the agent can recover; or a \textbf{silent failure}, where the action takes effect at the wrong target and no signal is raised. Because the verifier may also abstain, we trace a safety/applicability frontier rather than a single point. The same structure governs both modalities: a cheap check, an oracle-exact core, and the freedom to abstain, pointing to a general recipe for action-level oversight. To our knowledge, this is the first work to measure silent action failure under a construction that fixes ground truth before execution, and across two distinct agent modalities under one taxonomy.

\paragraph{Contributions.}
(1) A common framework for \emph{pre-action verification}: construction-defined ground truth, a success / clean-failure / silent-failure taxonomy, and abstention-based operating points (Sec.~\ref{sec:framework}).
(2) For \emph{shell commands}, a benchmark of 9{,}930 commands over 482 real tools with an anti-circular construction, and a static verifier whose error decomposition cleanly separates an oracle-exact, zero-false-positive core from a coverage-bounded flag check (Sec.~\ref{sec:cmd}).
(3) For \emph{code edits}, a benchmark of 640 edits over 224 files isolating the apply step, revealing a sharp content- vs.\ location-anchored safety dichotomy and a silent-failure mode that ordinary error handling misses (Sec.~\ref{sec:edit}).
(4) Two deployable guards, selective grounding with a two-tier gate and Robust-Apply (an anchor-and-verify meta-applier), that drive silent failure toward zero at a tunable cost, together with a cross-domain synthesis showing that one structure underlies both (Sec.~\ref{sec:synthesis}).

\paragraph{Relevance to AI Alignment.}
This is an \emph{empirical robustness evaluation} of LLM-agent actions and a \emph{safe-by-design} oversight mechanism: it quantifies a silent failure mode that ordinary error handling misses, isolates exactly where errors and false alarms originate, and supplies cheap guards that convert unrecoverable silent errors into recoverable clean failures. Both studies release open benchmarks, reproducible code, and practical evaluation tools, which the track explicitly encourages.

\section{Pre-Action Verification: A Common Framework}
\label{sec:framework}
An agent emits an action $a$ intended to produce an effect on a state $s$. The intended effect defines a target $T$. A deterministic verifier $V$ inspects $a$ (and $s$) before execution and either admits or refuses it. If the action is admitted, an executor produces a realized effect $R$.

\paragraph{Construction-defined ground truth.}
The central design choice is that $T$ is fixed \emph{before} any executor runs. For commands, validity is established by behavioral oracles that are independent of the verifier; for edits, the post-edit file is synthesized directly, then rendered into a format and perturbed. Because $T$ is known independently, every trial has a well-defined outcome:
\begin{itemize}\setlength{\itemsep}{1pt}
\item \textbf{Success}: $R$ realizes $T$.
\item \textbf{Clean failure}: the action is refused or does not take effect. \emph{Recoverable}: the agent is told and can retry or repair.
\item \textbf{Silent failure}: the action takes effect but $R\neq T$. \emph{Dangerous}: no signal is raised, so no recovery is triggered.
\end{itemize}
This taxonomy separates a true detection from a guess, and it makes safety and applicability independent axes. A verifier that refuses everything is trivially safe but useless; one that admits everything maximizes applicability but prevents nothing. What we study is the frontier between these extremes. Because silent failure is defined against a known target rather than inferred from a downstream symptom, a single scalar, the silent rate among admitted actions, captures the safety-relevant axis directly. That is what lets the two modalities be compared on the same footing.

\paragraph{Abstention and operating points.}
A verifier need not rule on every action. When the evidence is ambiguous it may abstain, admitting without a strong claim, or in a deployed gate raise a non-blocking warning. Sweeping the abstention rule traces a curve in (applicability, safety) space, and we report operating points along it. Each threshold $\theta$ induces a point $(\mathrm{app}(\theta),\,\mathrm{safe}(\theta))$, where applicability is the fraction of actions admitted and correct and safety is one minus the silent rate among admitted actions; the deployed guard maximizes applicability subject to a safety floor. The same structure recurs in both modalities. The cheap checks split into an oracle-exact part, definitively wrong when it fails and so yielding zero false positives, and a coverage- or guess-bounded part, a comparison against an imperfect reference that buys the remaining applicability and is where every false positive originates. Separating the two lets a deployment dial safety against applicability deliberately rather than accept one conflated number.

\paragraph{A worked example.}
Take the command \texttt{grep -rn -{}-colour pat .}\ The verifier resolves \texttt{grep} (binary check), confirms that it parses (\texttt{bash -n}), and tests each option against \texttt{grep}'s extracted flag set. Here \texttt{-{}-colour} is a genuine British-spelled alias that the parser may not have captured. A naive flag check would reject this valid command, giving a false positive, while an abstaining check withholds judgment and lets it through. An edit shows the same pattern. ``Insert after line 12'' carries no content to validate against, so a stale line number applies silently. The same change written as a search/replace block carries its surrounding anchor, so a stale anchor simply fails to match and yields a recoverable clean failure. In both modalities it is the representation, not the model, that decides whether a mistake can be caught before it lands.

\paragraph{Why before, not after.}
Post-hoc checks such as tests, a reviewing model, or a human in the loop are valuable but expensive, and they are blind to silent failure: a test passes if it does not cover the corrupted region, a reviewing model may miss a subtly wrong edit, and a human cannot audit every action of a fast agent. A pre-action check sidesteps this by refusing to let an unverifiable action take effect at all. The cost is occasional over-refusal of valid actions, which is recoverable, and in exchange the unrecoverable case is eliminated. Trading recoverable false alarms for the prevention of silent corruption is the design stance both guards share.

\paragraph{Threat model and scope.}
The verifier sees the action text and the local environment (installed tools, the current file). It does not see the model's intent and does not judge whether the intended change is correct, only whether the action as written can be realized validly and unambiguously. It is an oversight layer on action execution, separate from and composable with checks on action selection. We assume a non-adversarial environment (a buggy or hallucinating agent, not a malicious one) and a faithful executor; the guard targets unintended failure, not an agent that games the check or an executor that bypasses it, which call for a sandbox. Under these assumptions, the only way an error reaches the state is through an admitted action whose realized effect differs from its target, exactly the silent-failure case the framework exposes.

\section{Grounding Shell Commands}
\label{sec:cmd}
\paragraph{Benchmark.}
We build 9{,}930 commands over 482 installed tools. The valid commands ($n{=}1{,}986$) are drawn from the human-curated \texttt{tldr} corpus \citep{tldr}. The invalid commands, $1{,}986$ in each of four categories, are produced by mutation and confirmed invalid by behavioral oracles independent of the verifier: a nonexistent binary (resolved by \texttt{which}), malformed syntax (rejected by \texttt{bash -n}), an invalid long option, and an invalid short option (both absent from a flag set extracted from \texttt{-{}-help}, \texttt{-h}, and \texttt{man}). This anti-circular construction keeps the verifier from being graded against its own assumptions: a benchmark generated by the rules the verifier checks would measure self-consistency, not detection, so our oracles act as independent witnesses of (in)validity. The construction funnel is explicit. Of 6{,}474 \texttt{tldr} tool pages, 546 tools are installed; parsing yields 2{,}160 candidate commands, or 2{,}145 once every pipeline stage must resolve, capped per tool to a final 1{,}986.

\paragraph{Verifier.}
The verifier runs three checks: syntax (\texttt{bash -n} on the command string), binary existence (\texttt{which}), and flag validity (membership in the extracted flag set). It is subcommand-aware, so that \texttt{git commit} flags are treated separately from \texttt{git} flags. When a tool yields no extractable flags, the verifier abstains and admits the command, so flag recall is bounded by extraction coverage (91.1\% of tools yield a non-empty set); we never guess against an empty reference. Each check is a pure function of the command and the environment. \texttt{bash -n} is the shell's own parser, \texttt{which} consults the live \texttt{PATH}, and the flag set is built once per tool by parsing \texttt{-{}-help}, \texttt{-h}, and \texttt{man} into long and short options. Because the verifier inherits the shell's notion of well-formedness rather than re-implementing it, its structural checks are exact rather than heuristic.

\paragraph{Results.}
The full configuration detects \textbf{95.8\%} of invalid commands at a \textbf{10.0\%} false-positive rate ($P{=}0.975$, $F_1{=}0.966$). The estimates are stable under a tool-clustered bootstrap (95\% CIs: recall $[0.944,0.971]$, FPR $[0.078,0.122]$, $B{=}2000$ over 482 tool clusters), and the verifier is cheap: a warm median of 2.55\,ms per command (p95 2.78\,ms, p99 3.05\,ms), with a one-time cold cost dominated by flag extraction (p95 2.85\,s, cached thereafter). The decomposition is the scientific point (Table~\ref{tab:cmd}). The syntax and binary checks are oracle-exact: \texttt{bash -n} and \texttt{which} test exactly the property that defines a malformed or unknown-binary command, so they cannot reject a valid command (\textbf{zero} false positives) yet still catch half of all errors (recall $0.500$). The binary-only check is the naive ``does \texttt{command -v} resolve?'' baseline (recall $0.255$). The flag check supplies the rest of the recall and accounts for every false positive: all 198 are flags the tool accepts but our parser failed to extract, falling into three token classes (48.5\% single-dash single-character, 29.8\% single-dash multi-character, 21.7\% double-dash) and spread thinly over 95 tools, at most 8 for any one. False positives are a property of help-text quality, not of the model or of the idea of grounding. By category, the oracle-exact checks catch nonexistent-binary and malformed-syntax errors perfectly (recall $1.000$ each), while invalid long and short flags are caught at $0.916$ and $0.916$, bounded by the $91.1\%$ flag-extraction coverage rather than by the method (Table~\ref{tab:percat}).

\begin{table}[t]
\centering

\small
\setlength{\tabcolsep}{4pt}
\begin{tabular}{lcccc}
\toprule
\textbf{Configuration} & \textbf{Prec.} & \textbf{Recall} & \textbf{$F_1$} & \textbf{FPR} \\
\midrule
Full (subcmd-aware)          & 0.975 & 0.958 & 0.966 & 0.100 \\
Conservative (long-only)     & 0.992 & 0.735 & 0.844 & \textbf{0.023} \\
\textbf{Selective (ours)}    & 0.982 & 0.958 & \textbf{0.970} & 0.070 \\
\midrule
Syntax + binary              & 1.000 & 0.500 & 0.667 & \textbf{0.000} \\
Binary only (\texttt{command~-v}) & 1.000 & 0.255 & 0.407 & \textbf{0.000} \\
\bottomrule
\end{tabular}

\caption{Shell-command verifier: configurations and key ablations ($N{=}9{,}930$). FPR is on the 1{,}986 valid commands. Syntax+binary form an oracle-exact, zero-FP core; the flag check is the only source of false positives.}
\label{tab:cmd}
\end{table}

\begin{figure}[t]
\centering
\includegraphics[width=0.92\columnwidth]{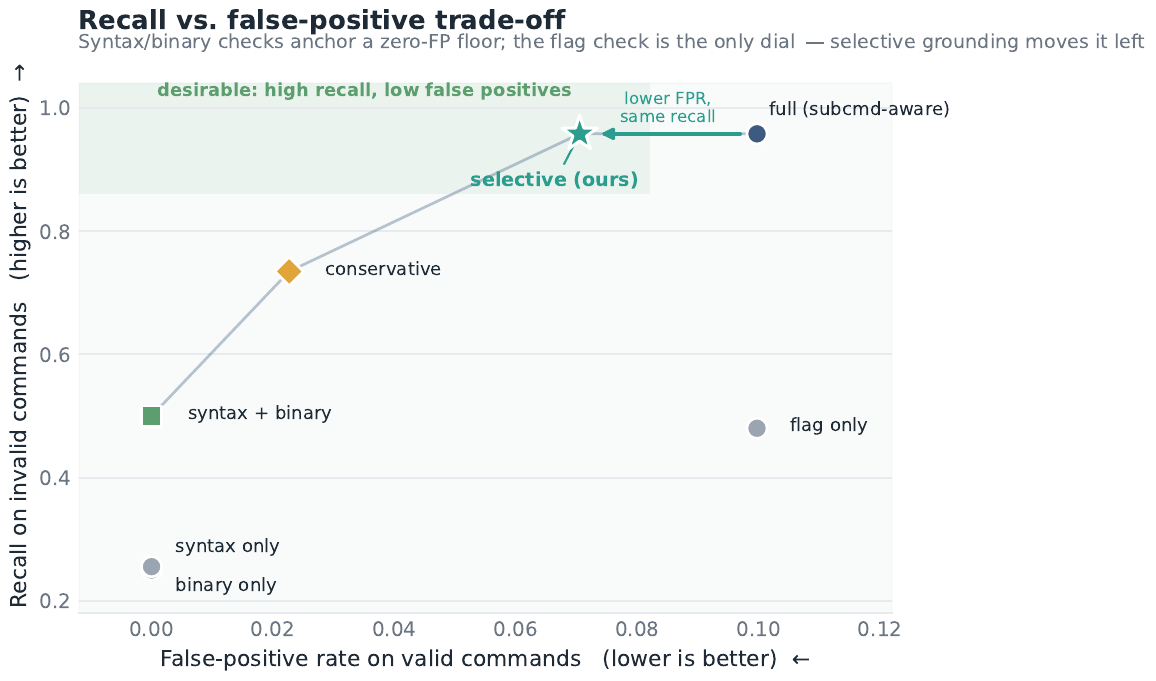}
\caption{Recall vs.\ false-positive trade-off for the shell-command verifier. Syntax and binary checks anchor a zero-false-positive floor (catching half of all errors); the flag check is the only dial. Selective grounding (ours) moves the operating point left, giving lower false positives at equal recall, and Pareto-dominates the naive full configuration.}
\label{fig:cmd}
\end{figure}

\paragraph{Selective grounding and a two-tier gate.}
Every false positive is an extraction gap, and the most ambiguous token is the single-dash multi-character option, a BSD-style cluster that may or may not be a real flag, so we abstain on exactly those tokens. This is lossless by construction: none of the injected invalid errors is a single-dash multi-character token, so abstaining can only remove false positives. It also Pareto-dominates the naive policy, giving the same \textbf{0.958} recall at \textbf{7.0\%} false positives rather than 10.0\% (Figure~\ref{fig:cmd}). Deployed as a gate, the oracle-exact checks block (recall $0.500$ at $0\%$ false positives), while flag suspicions become non-blocking warnings (a 7.0\% warn-rate on valid commands). A wrong block is then impossible, and the agent always receives a recoverable signal it can act on.

\paragraph{Where the false alarms live.}
Table~\ref{tab:fp} breaks the 198 false positives down by token class. Selective grounding abstains on the single-dash multi-character class, which is 29.8\% of false positives and the BSD-style cluster whose flag membership is genuinely ambiguous from help text. That is why abstaining removes three points of false-positive rate, from 10.0\% to 7.0\%, at no cost in recall. The remaining false positives are extraction gaps that a richer help-text parser or a per-tool completion spec would close. None reflects a limitation of grounding itself, and all are confined to 95 of 482 tools, at most 8 for any one.

\begin{table}[t]
\centering

\small
\setlength{\tabcolsep}{4pt}
\begin{tabular}{lcc}
\toprule
\textbf{Token class} & \textbf{Share of FPs} & \textbf{Abstained?} \\
\midrule
Single-dash, one char    & 48.5\% & no \\
Single-dash, $>$one char & 29.8\% & \textbf{yes} \\
Double-dash (long)       & 21.7\% & no \\
\bottomrule
\end{tabular}

\caption{False-positive composition: all 198 are real flags missed by extraction. Selective grounding abstains on the ambiguous single-dash multi-character class.}
\label{tab:fp}
\end{table}

\begin{table}[t]
\centering
\small
\setlength{\tabcolsep}{4pt}
\begin{tabular}{lcc}
\toprule
\textbf{Invalid category} & \textbf{Recall} & \textbf{95\% CI} \\
\midrule
Nonexistent binary  & 1.000 & $[0.998,\,1.000]$ \\
Malformed syntax    & 1.000 & $[0.998,\,1.000]$ \\
Invalid long flag   & 0.916 & $[0.903,\,0.928]$ \\
Invalid short flag  & 0.916 & $[0.903,\,0.927]$ \\
\bottomrule
\end{tabular}
\caption{Recall by error category. Oracle-exact checks (binary, syntax) catch structural errors perfectly; flag errors are caught at rates bounded by help-text extraction coverage, not by the method.}
\label{tab:percat}
\end{table}

\section{Applying Code Edits}
\label{sec:edit}
\paragraph{Benchmark.}
We isolate the apply step, which is logically distinct from generating the right change. We take 640 edits over 224 real Python files. Each edit's post-edit target is synthesized first, then rendered into one of four formats and perturbed only on its location surface. Because the target is fixed before any applier runs, correctness does not depend on the applier and silent misapplication is directly observable. This inverts the usual setup. Instead of asking whether an applier reproduces an unknown intended file, we define the intended file and ask only whether the applier reaches it, so a wrong location is unambiguously a silent failure rather than a disagreement about intent. The formats are search/replace (exact, whitespace-normalized, and fuzzy at threshold $\tau$), unified diff (real GNU \texttt{patch} \citep{gnudiff} at fuzz factors 0 and 2), line ranges, and whole-function-by-name. The perturbations include reindentation and line shifts, giving $23{,}040$ trials.

\paragraph{Formats and perturbations.}
The four formats span how agents address a change. Search/replace quotes the code to be changed (content-anchored); unified diff quotes context lines around it (content-anchored, validated independently by \texttt{patch}); line ranges name positions and whole-function edits name the function (both location-anchored). The perturbations model realistic drift between when an agent reads a file and when its edit lands: reindentation, which changes whitespace but leaves content intact, and line shifts, which insert or delete lines above the target. A safe applier fails cleanly under drift it cannot resolve; a dangerous one applies anyway, to the wrong place.

\begin{figure}[t]
\centering
\includegraphics[width=0.92\columnwidth]{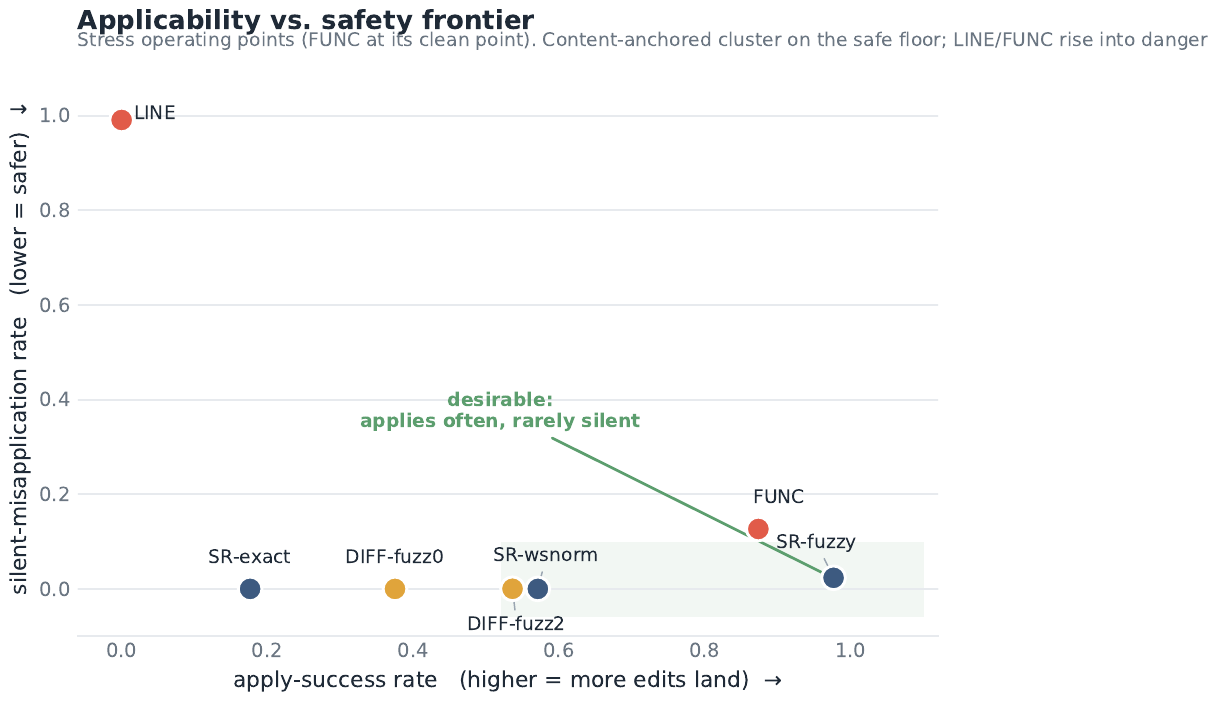}
\caption{Applicability vs.\ safety for edit formats. Content-anchored formats occupy the safe floor (never silent), trading applicability for safety as matching loosens; location-anchored formats are pushed up into the dangerous region. Whole-function is shown at its clean operating point.}
\label{fig:edit}
\end{figure}

\begin{figure}[t]
\centering
\includegraphics[width=0.96\columnwidth]{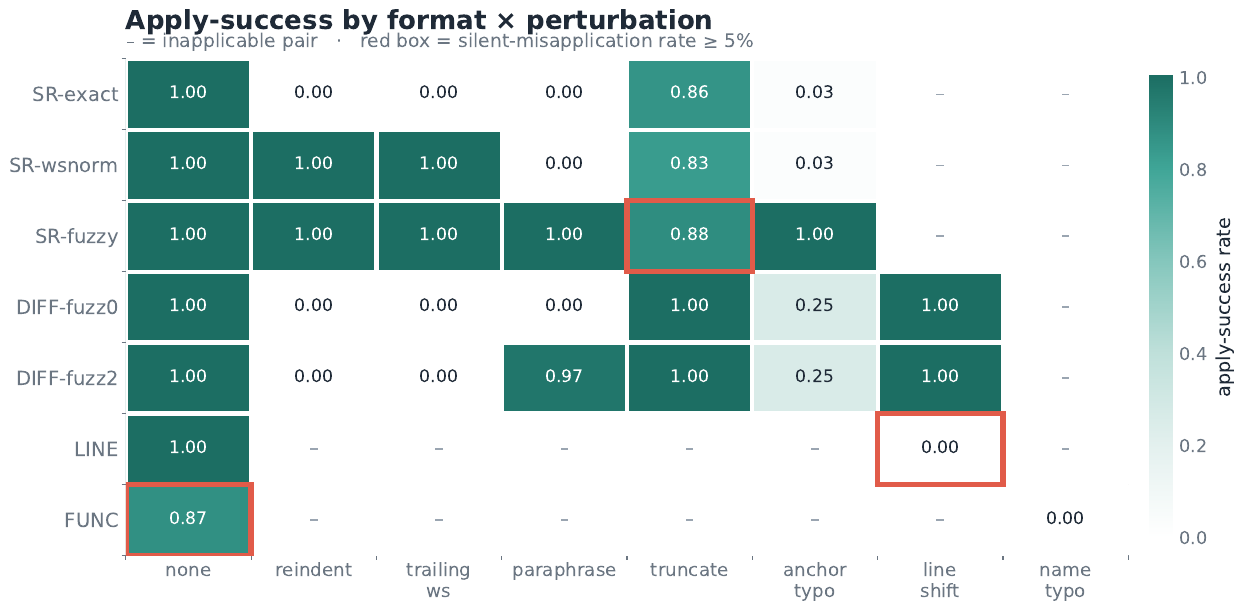}
\caption{Silent-misapplication rate by format (rows) and perturbation (columns). Content-anchored formats stay at the safe floor across all perturbations; location-anchored formats light up: line ranges under shifts, and whole-function from name collisions even with no perturbation.}
\label{fig:heat}
\end{figure}

\paragraph{A safety dichotomy.}
The formats split cleanly (Table~\ref{tab:edit}, Figure~\ref{fig:edit}). Content-anchored formats locate the edit by the surrounding code and fail cleanly when they cannot match: search/replace and unified diff never misapply silently (0.000), trading applicability for safety as matching loosens (SR-fuzzy stress success $0.977$, cluster CI $[0.972,0.981]$). Location-anchored formats locate by position or name and fail silently: a one-line shift leaves line ranges applying nothing correctly while \textbf{silently corrupting 99.1\%} of files, and whole-function edits hit the wrong same-named function \textbf{12.7\%} of the time even unperturbed, from duplicate-name collisions. The heatmap (Figure~\ref{fig:heat}) shows the effect is uniform across perturbations, not an artifact of one operator.

\begin{table}[t]
\centering

\small
\setlength{\tabcolsep}{4pt}
\begin{tabular}{lccc}
\toprule
\textbf{Format} & \textbf{Clean succ.} & \textbf{Stress succ.} & \textbf{Stress silent} \\
\midrule
SR-exact            & 1.000 & 0.176 & \textbf{0.000} \\
SR-wsnorm           & 1.000 & 0.571 & \textbf{0.000} \\
SR-fuzzy            & 1.000 & 0.977 & 0.023 \\
Unified diff        & 1.000 & 0.375--0.536 & \textbf{0.000} \\
Line range          & 1.000 & 0.000 & \textbf{0.991} \\
Whole function      & 0.873 & 0.000 & 0.127$^{\dagger}$ \\
\bottomrule
\end{tabular}
\\[2pt]
{\footnotesize $^{\dagger}$ Whole-function silent failure occurs \emph{unperturbed}, from duplicate-name collisions; unified diff uses real GNU \texttt{patch}.}

\caption{Edit-application outcomes by format. ``Clean'' = no perturbation; ``Stress'' pools applicable perturbations. Lower silent is safer; content-anchored formats are silent-free.}
\label{tab:edit}
\end{table}

\paragraph{Safety is format and policy.}
A zero silent rate is a property of the applier policy as much as of the format. Re-running the content-anchored conditions with a permissive first-match policy raises silent misapplication from $0.000$ to $0.020$ (exact) and $0.023$ (whitespace-normalized). The format still dominates: content-anchored silent rates stay near 2\% even when permissive, against 99\% for line numbers. But reaching zero requires a refuse-ambiguous policy. A benchmark that reports only one of format or policy conflates the two.

\paragraph{Multi-hunk and generalization.}
Fuzzy matching is a tunable dial. It is safe on single edits (2.3\% silent) but dangerous on multi-hunk edits, where shrinking per-hunk anchors push its silent rate to \textbf{39\%}. It also never abstains there (clean-failure rate $0.000$), so every wrong location is silent rather than recoverable. Unified diff is unaffected, because each hunk carries its own context and \texttt{patch} validates that context independently. The dichotomy and the rates replicate on a third-party codebase, 55 edits from the \texttt{requests} library: SR-fuzzy's silent rate is $0.025$ (against $0.023$ on the standard library) and line ranges stay catastrophic ($1.000$ silent), which indicates that the effect is intrinsic to the formats rather than to one corpus. Unified diff is the robust outlier. At fuzz 0 it demands exact context, and at fuzz 2 it tolerates small drift, but in neither case does it commit without matching context, so it stays silent-free even across two hunks. The cost is the lowest applicability among content-anchored formats under heavy reindentation.

\begin{table}[t]
\centering

\small
\setlength{\tabcolsep}{4pt}
\begin{tabular}{lcc}
\toprule
\textbf{Format} & \textbf{Single-hunk silent} & \textbf{Multi-hunk silent} \\
\midrule
SR-exact      & 0.000 & 0.000 \\
SR-wsnorm     & 0.000 & 0.000 \\
SR-fuzzy      & 0.023 & \textbf{0.393} \\
Unified diff  & 0.000 & 0.000 \\
Line range    & 0.991 & 1.000 \\
\bottomrule
\end{tabular}

\caption{Multi-hunk edits: fuzzy matching's silent rate compounds as per-hunk anchors shrink, while exact, whitespace-normalized, and diff matching stay safe and line ranges remain catastrophic.}
\label{tab:mh}
\end{table}

\paragraph{Robust-Apply.}
The remedy mirrors selective grounding: verify, and refuse when unsure (Algorithm~\ref{alg:robust}). Robust-Apply is a format-agnostic meta-applier. It requires a minimum anchor size $K_{\min}$, refuses ambiguous matches, and accepts a fuzzy match only when it clears a similarity floor $\tau_{hi}$ and beats the runner-up by a margin $\delta$, committing through real \texttt{patch}. With $(K_{\min},\tau_{hi},\delta){=}(2,0.90,0.20)$ it records \textbf{one} silent misapplication in \textbf{8{,}320} trials ($0.01\%$), across every format, every perturbation, the multi-hunk edits, and the third-party codebase, against native rates as high as $99.1\%$. On the trials where aggressive fuzzy misapplies silently, it converts \textbf{100\%} to recoverable clean failures while preserving 60\% of the applies that fuzzy achieved, and it is lossless on clean content-anchored edits. The price is applicability ($0.51$ against $0.98$ on perturbed single-hunk edits), an explicit and tunable trade. The result is not knife-edge. Across the swept range $\tau_{hi}\in[0.85,0.95]$ and $\delta\in[0.10,0.30]$, the silent rate stays at or below $0.05\%$ while applicability varies smoothly, so the operating point can be tuned to a deployment's tolerance without re-introducing silent failure. The one silent case among the 8{,}320 trials occurred under unified-diff stress, where GNU \texttt{patch} committed a hunk at a shifted location at nonzero fuzz (rate $0.0005$); constraining the fuzz factor removes it, so even the residual risk is a tunable knob rather than a floor.

\begin{algorithm}[t]
\caption{Robust-Apply (anchor-and-verify)}
\label{alg:robust}
\textbf{Input}: file $F$; edit with anchor $A$, replacement; params $K_{\min},\tau_{hi},\delta$\\
\textbf{Output}: edited file, or \textsc{Clean-Fail} (recoverable)
\begin{algorithmic}[1]
\IF{$|A| < K_{\min}$}
  \STATE \textbf{return} \textsc{Clean-Fail} \COMMENT{anchor too small to trust}
\ENDIF
\STATE try exact, then whitespace-normalized match of $A$ in $F$
\IF{a \emph{unique} such match $m$ exists}
  \STATE \textbf{return} apply replacement at $m$
\ENDIF
\STATE $(m_1,s_1),(m_2,s_2)\gets$ top-2 fuzzy matches by similarity
\IF{$s_1 \ge \tau_{hi}$ \textbf{and} $s_1-s_2 \ge \delta$}
  \STATE \textbf{return} apply at $m_1$ via \texttt{patch}
\ENDIF
\STATE \textbf{return} \textsc{Clean-Fail} \COMMENT{ambiguous; refuse rather than guess}
\end{algorithmic}
\end{algorithm}

\paragraph{Which operating point to use.}
The choice is a deployment decision. Aggressive fuzzy is the right pick when applicability is paramount and a downstream check (tests, review, or self-repair) will catch a wrong edit; its $2.3\%$ single-hunk silent rate is then a cost the check recovers. Robust-Apply is preferable when a silent wrong edit is unacceptable, or when no reliable downstream check exists. Exact or whitespace-normalized matching is the simplest zero-tuning safe default. The one combination to avoid is aggressive fuzzy with no downstream check, where silent corruption accumulates unobserved.

\section{Cross-Domain Synthesis}
\label{sec:synthesis}
The two studies are instances of one structure (Table~\ref{tab:synth}). In each, a cheap deterministic check before the action takes effect, together with the freedom to abstain, converts silent failures into recoverable clean failures at a tunable cost in applicability. In each, the checks split into an oracle-exact part that gives a zero-false-positive safety floor (syntax and binary for commands, exact and whitespace-normalized matching for edits) and a coverage- or guess-bounded part that buys the remaining applicability and is the only source of error (flag extraction, fuzzy matching). And in each, the dangerous representations are exactly those that cannot be validated against content: a flag absent from any reference, or a line number or function name with no surrounding anchor. Such representations cannot tell a right target from a wrong one, and so they fail silently.

\begin{table}[t]
\centering

\small
\setlength{\tabcolsep}{3.5pt}
\begin{tabular}{p{0.215\columnwidth}p{0.34\columnwidth}p{0.34\columnwidth}}
\toprule
 & \textbf{Shell commands} & \textbf{Code edits} \\
\midrule
Ground truth     & behavioral oracles            & synthesized target \\
0-FP core        & syntax + binary               & exact / ws-norm match \\
Error source     & flag-extraction coverage      & fuzzy threshold \\
Unsafe form      & unknown flag                  & line \#\ / function name \\
Guard            & selective two-tier gate       & Robust-Apply \\
Operating pt.    & R 0.958 @ FPR 0.07            & 1 silent / 8{,}320 \\
\bottomrule
\end{tabular}

\caption{One structure, two modalities. A cheap check with an oracle-exact core and the option to abstain converts silent failures into recoverable ones in both settings.}
\label{tab:synth}
\end{table}

For agent oversight, this suggests a concrete and model-agnostic principle. Prefer action representations that can be checked against content, gate them with an oracle-exact blocker together with an abstaining soft check, and treat applicability as a dial set by how much recoverable failure a deployment can tolerate.

\paragraph{Beyond two modalities.}
The recipe extends to other agent actions with a checkable surface. A file operation can confirm a path exists and is writable before acting; an API call can validate its arguments against a schema and refuse on ambiguity; a staged database write can be checked against a constraint before commit. The design question is always the same: is there an oracle-exact property that is definitively wrong when violated, and a content anchor the action can be matched against? Where both exist, a cheap pre-action guard applies; where neither does, the action interface itself is what to reconsider. A small library exposing the success / clean-failure / silent-failure contract and an \texttt{abstain} primitive, into which per-modality verifiers plug, would turn this into reusable infrastructure. Because the guards cost microseconds to milliseconds and no model call, they work as always-on wrappers around frontier agents whose internals are fixed or unavailable, complementing scalable-oversight methods that operate on the model itself.

Two properties make this appealing as oversight. First, the guard's failures are recoverable by design: a refused action returns a clean signal the agent can retry or repair, so a false alarm costs a wasted step rather than a corrupted state. Second, the guard is auditable: its decisions are deterministic functions of the action and the environment, not opaque model judgments, so a reviewer can see exactly why an action was blocked or warned. A learned critic would add inference cost, latency, and a fresh opportunity for silent error; a deterministic pre-action check adds none of these, and it can wrap a learned proposer as the final, verified commit step. This is a small but concrete instance of safe-by-design engineering: the safety property, no silent misapplication, is established by construction and measured rather than hoped for. The oracle-exact core admits a small provable guarantee: \texttt{bash -n}, \texttt{which}, and exact matching decide the property that defines (in)validity, so the zero-false-positive floor holds by construction. This is short of a full formal safety case, since the soft checks remain empirical, but it is a verified core beneath the abstaining layer; we state the guarantees precisely as three propositions in the supplementary material.

\paragraph{Where the principle stops.}
Pre-action verification addresses whether an action can be realized validly and unambiguously, not whether it is semantically right: a valid command with a harmful effect, or an edit applied to the right place but encoding a logic error, lies outside its reach. The guards also presume a checkable structure; an action with no content to anchor against, such as a raw byte offset or an opaque handle, cannot be grounded and is better discouraged at the interface. Our contribution is to make the realizability layer safe and cheap, freeing more expensive oversight (tests, review, learned critics) to concentrate on semantic correctness. The clearest extension is to measure these guards in the loop: running several frontier agents over real trajectories to estimate how often the silent cases we construct actually arise and how much the guards lift end-to-end reliability, which our single-model probes only begin to provide.

\paragraph{Versus other guards.}
Two common alternatives are worth comparing against: sandboxing the executor and rolling back on error, and asking a second model to review each action. Pre-action verification is cheaper than the sandbox, needing no snapshot or rollback machinery and catching the silent case a rollback never triggers on, and more predictable than the reviewer, adding no inference and no fresh hallucination. It is also narrower, guarding realizability rather than semantics. The three are complementary, and the cheapest, most auditable layer, ours, is the natural first line. Rollback in particular cannot catch a silent failure, since it fires only on an observed error and a silent misapplication raises none; differential testing or a second-model reviewer then supplies the semantic check verification omits. Combining them is defense in depth, each layer covering a class of error the others miss.

\paragraph{Deployment cost.}
Both guards run inline on every action with no model call, GPU, or network. The command verifier adds a median of 2.55\,ms once a tool's flags are cached, apart from a one-time cold extraction, and Robust-Apply is dominated by a single similarity scan. On abstention, each returns a structured signal the agent can act on, such as the suspected flag or the ambiguous match and its runner-up, rather than a bare rejection.

\section{Threats to Validity}
\textbf{Construct.} The invalid commands and perturbed edits are synthetic, confirmed by oracles or fixed by construction. They need not match the distribution of real model errors, though the failure mechanisms, extraction gaps and location drift, are intrinsic to the representations. The zero-false-positive claim for syntax and binary checks is exact on complete commands but carries deployment caveats: \texttt{bash -n} rejects incomplete fragments, and \texttt{which} misses aliases, shell functions, and a session-updated \texttt{PATH}. A deployed gate should treat these cases as abstentions. \textbf{External.} We probe real model output only lightly (42 commands and 9 edits from an Anthropic Claude model, specific version not recorded, all valid), so this evidence bears on the false-positive side and does not estimate recall against a real error distribution. A multi-model study over agent trajectories is the key next step. The results come from one Linux host with Python and shell tooling, and exact rates need not transfer to other languages, larger functions, or non-function code, even where the qualitative dichotomies should hold. \textbf{Internal.} Latency is the only non-deterministic measurement; every detection metric is deterministic given the committed snapshots and fixed seeds. Extended results (per-category recall, the full $\tau$ and line-shift sweeps, the multi-hunk and format-versus-policy tables, cluster intervals, and the agent traces) are in the supplementary material.

\section{Related Work}
Agent benchmarks measure end-to-end task success \citep{swebench,agentbench,intercode,osworld} but rarely isolate why an action fails; we target the action-validity and apply steps directly, with construction-defined ground truth that makes silent failure observable. Tool- and API-use work studies whether models select and call the right function \citep{gorilla,toolllm,toolformer,react}, and NL-to-command generation whether they produce correct flags and structure \citep{nl2bash}; we reuse that difficulty as a verification signal rather than a generation target. Self-refinement and self-repair correct outputs after an error is observed \citep{selfrefine,selfrepair}; pre-action verification is complementary, cheaper, and addresses the silent case no observation would flag. General shell linters such as ShellCheck target script-level correctness rather than flag validity against the installed tool, and completion frameworks such as argcomplete operate at the prompt rather than as an agent gate; neither grounds a command before execution. Edit formats are a practical concern across coding-agent toolchains \citep{aider,agentless,sweagent,openhands}, where matching builds on standard diff and similarity methods \citep{myers1986,ratcliff1988,gnudiff}; we give the first controlled measurement of their silent misapplication. Execution-safety work sandboxes or gates agent actions to contain damage \citep{intercode,osworld}; our gate acts earlier, refusing an ill-formed action before any effect rather than containing its aftermath. Closest in spirit is work that verifies or tests generated artifacts before use \citep{selfrepair,agentless}, but that line typically runs the artifact and inspects the result, whereas we verify realizability statically, which is what makes the silent case observable and preventable. Our success / clean-failure / silent-failure taxonomy is deliberately operational, keyed to agent recoverability rather than a general bug taxonomy, which lets a single number, the silent rate, capture the safety-relevant axis. More broadly, ours is a narrow, deterministic instance of the scalable-oversight goal of keeping agent behaviour checkable as capability grows \citep{amodei2016concrete,bowman2022scalable,irving2018debate}, and it complements formal verification of learned components such as SMT-based neural-network checking \citep{katz2017reluplex}: we verify an action's realizability against the environment, not properties of the model. Confidence intervals use the method of \citet{wilson}.

\section{Conclusion}
A wrong agent action that fails silently is an oversight failure that ordinary error handling misses. Studying shell commands and code edits, we found that fixing an action's correct effect by construction turns silent failure from an inferred symptom into a measured quantity. With that in place a consistent picture emerges: the cheap checks separate into an oracle-exact core that raises no false alarm and a softer, coverage-bounded remainder that carries all the error, and a representation is safe exactly when the action can be checked against its content. The resulting guards are model-agnostic, run in microseconds to milliseconds, and ship with the benchmarks, so they can sit in front of a deployed agent as it acts. The wider lesson is about interface design: an action checkable against content can be made safe cheaply, while one that cannot leaves a silent failure that no downstream test, review, or repair fully recovers. That is the case for building agent interfaces around checkable actions from the start, and the reason we expect the construction-defined method to reach well beyond the two modalities studied here.

\section*{Ethical Statement}
This work is defensive. The guards prevent silent corruption of files and unintended command execution, and their failure mode, a refused action, wastes a step rather than damaging state. The benchmarks use only public tools and source code, with no human subjects or personal data. The dual-use surface is small: knowing that line-number edits misapply silently helps a defender choose safe formats far more than it helps an attacker, so releasing the benchmarks, guards, and findings improves agent safety more than it aids misuse.

\bibliography{references}

\end{document}